\documentclass[11pt,a4paper]{article}
\usepackage[hyperref]{eacl2021}
\usepackage{times}
\usepackage{graphicx}
\usepackage{latexsym}

\usepackage{microtype}

\aclfinalcopy 

\title{Improved Customer Transaction Classification using Semi-Supervised Knowledge Distillation }

\author{Rohan Sukumaran \textsuperscript{$^*$} \\
  Applied Research, Swiggy \\
  \texttt{rohan.s16@iiits.in}
  }

\date{}

\begin{document}
\maketitle
\renewcommand{\thefootnote}{\fnsymbol{footnote}}
\footnotetext[1]{Work done while an intern at Applied Research, Swiggy.}
\renewcommand*{\thefootnote}{\arabic{footnote}}
\begin{abstract}
In pickup and delivery services, transaction classification based on customer provided free text is a challenging problem. It involves association of a wide variety of customer inputs to a fixed set of categories while adapting to the various customer writing styles. This categorization is important for the business: it helps understand the market needs and trends, and also assist in building a personalized experience for different segments of  the customers. Hence, it is vital to capture these category information trends at scale, with high precision and recall. In this paper, we focus on a specific use-case where a single category drives each transaction. We propose a cost-effective transaction classification approach based on semi-supervision and knowledge distillation frameworks. The approach identifies the category of a transaction using free text input given by the customer. We use weak labelling and notice that the performance gains are similar to that of using human annotated samples. On a large internal dataset and on  20Newsgroup dataset, we see that RoBERTa performs the best for the categorization tasks. Further, using an ALBERT model (it has 33x less parameters vis-a-vis paremeters of RoBERTa), with RoBERTa as the Teacher, we see a performance similar to  that of RoBERTa and better performance over unadapted ALBERT. This framework, with ALBERT as student and RoBERTa as teacher, is further referred to as R-ALBERT in this paper. The model is in production and is used by business to understand changing trends and take appropriate decisions.
\end{abstract}

\section{Introduction}
Natural Language Understanding (NLU) has become a ubiquitous part of the e-commerce platforms with applications in information discovery~\cite{degenhardt2019ecom,grandhi2016methods}, sentiment analysis~\cite{liang2019linguistic, taylor2020commerce}, query understanding~\cite{zhao2019dynamic} and many other similar tasks. The advancement in machine learning has enhanced the scale of adaption of such capabilities to enable richer customer experience. 

In this paper, we consider a use-case of pickup and delivery services where customers make use of short sentences or lists to inform about the products to be picked up (or bought) from a certain location and dropped at a target location. Table~\ref{table:examples} shows a few examples of the descriptions used by our customers to describe their products. Customers tend to use short, code-mixed (using more than one language in the same message\footnote{In our case, Hindi written in roman case is mixed with English}) and incoherent textual descriptions of the products for describing them in the transaction. These descriptions, if mapped to a fixed set of categories, will help assist critical business decisions such as how to enhance the customers' experience on the platform, understand the importance of each category and issues faced by them, demographic driven prioritization of categories, launch of new product categories etc. As expected, a transaction may comprise of multiple products - this adds an additional level of complexity to the task. In this work, we focus on a multi-class classification of transactions, where a single majority category drives the transaction.

\begin{table}[bht]
\centering
\begin{tabular}{ |p{5cm}|p{2cm}| }
 \hline
 \textbf{Transaction Description} & \textbf{Category}\\
 \hline
 \textit{"Get me my lehanga"} \newline
 \textbf{Translation :} Get me my skirt &   Clothes \\
 \hline
\textit{"Buy a 500gm packet of Whole grain Atta"} & Grocery \\
 \hline
 \textit{"Get a roll of  paratha"} &   Food\\
 \hline
 \textit{"Mera do bags leaoo"} \newline
  \textbf{Translation :} Bring two of my bags &   Package  \\
 \hline
\end{tabular}
\caption{Samples of actual transaction descriptions used by our customers along with their corresponding categories as labelled by the subject matter experts (SMEs).}
\label{table:examples}
\end{table}

Classifying transaction types to the right categories requires labelled data for training an ML model. Some of the data used in this paper (Table~\ref{tab:my-table-data}: Train data) was labeled manually by the subject matter experts (SMEs) from the business and it was a very expensive exercise.

Recent works have shown that BERT based models~\cite{conneau2019unsupervised, yang2019xlnet} are able to achieve state-of-the-art performance on text classification tasks. Through experiments we could observe that RoBERTa~\cite{liu2019roberta} was the best performing model for our task. However, owing to the large number of parameters, it was not feasible to deploy this model at scale. Furthermore, we observed that the lighter versions of BERT such as ALBERT (base)~\cite{lan2019albert, wolf2019huggingface} that are production friendly could not match the performance of RoBERTa.

With the above two limitations for production friendliness in mind, we propose: a) an approach that leverages semi supervision to reduce the manual labeling cost and, b) also explore knowledge distillation to build a smaller model (in terms of numbers of parameters) that matches the performance of the state-of-the-art heavier models such as RoBERTa. The key contributions of this paper are:

\begin{itemize}
\item \textbf{Weak Labelling:} A framework based on semi-supervised learning and weak supervision to reduce manual data labelling bottleneck.

\item \textbf{Knowledge Distillation Framework:} Training a lightweight model (33x lesser parameters) with the help of weak labels, which is able to match the performance of a much heavier model.

\item \textbf{Inference and Scalability:} Attained 86 times increase in inference speed with the Student model while running on a r5 4x large (Intel Xeon Platinum 8000 series processor (Skylake-SP) with a sustained all core Turbo CPU clock speed of up to 3.1 GHz) machine in production when compared to that of the Teacher model.
\end{itemize}

\section{Related Work} \label{related}
Text classification problems with code-mixed inputs have been studied and transformer based models perform well on benchmarks~\cite{chang2019x, lu2020vgcn} like TREC-6~\cite{10.1016/S0306-4573(99)00043-6} and DBpedia~\cite{auer2007dbpedia}. The larger size of the models exacerbates the challenge of deployment  with limited resources~\cite{chen2020adabert, sajjad2020poor}. Multiple methods like  quantization~\cite{zafrir2019q8bert}, pruning~\cite{gordon2020compressing}, distillation~\cite{sanh2019distilbert, jiao2019tinybert} and weight sharing~\cite{houlsby2019parameter} are used to mitigate this issue. All these methods have shown varying degrees of success compared to the performance of the base model from which they are derived.  
\subsection{Weak labeling of data}
Data-less text classification has become a popular practice to achieve low cost model training. \cite{seedword2014} explored approaches based on topic modelling to predict labels for documents. Our problem setting involves short transaction descriptions that do not perform well with standard topic modelling techniques. \cite{pseudolabel2018} worked with unlabeled data by identifying a minimal set of seed word based pseudo labels for documents and trained a Naive Bayes model using semi-supervision. We had a large amount of manually tagged data - we leveraged this to extract candidate training samples from unlabeled data.  
\subsection{Knowledge Distillation} 
\cite{hinton2015distilling} studied how a model can be used to label unlabelled data and make use of the model predictions (the probability distributions and/or the one hot encoded labels) for training using a combination of loss functions. We borrow from~\cite{hinton2015distilling} the concept of using different loss functions during training. In this paper, we use transformer models as teacher and student. In~\cite{yuan2019revisit}, the authors proposed how a model could be it's own teacher and how lightweight models can teach a heavier model. Although this showed promising results in  image classification tasks, we observed that the similar performance gains was not visible for our setting of NLP tasks. This could be due to the difference in the underlying information captured in representation learning. Hence, we stick to the standard teacher-student framework where heavier models are used as the teacher.

\subsection{Fine-tuning on target tasks before teaching}
The Multifit~\cite{eisenschlos2019multifit} paper proposed to fine-tune a lightweight model on the target task with labelled or pseudo labelled data and empirically showed the performance improvement derived from using the teacher model. We leverage a similar setup and validate the generalization of the method to large scale transformer based models. We observe this to be an effective training strategy. The study also showed that the student model was robust to noise as the teacher served as a regularizer. In~\cite{tamborrino2020pre} also the authors explain the benefits of relevant knowledge transfer via task specific fine-tuning.
\section{Methodology}\label{Methodology}
\begin{figure}[ht]
\includegraphics[width=7.5cm, trim={0cm 0cm 0cm 0cm}, clip]{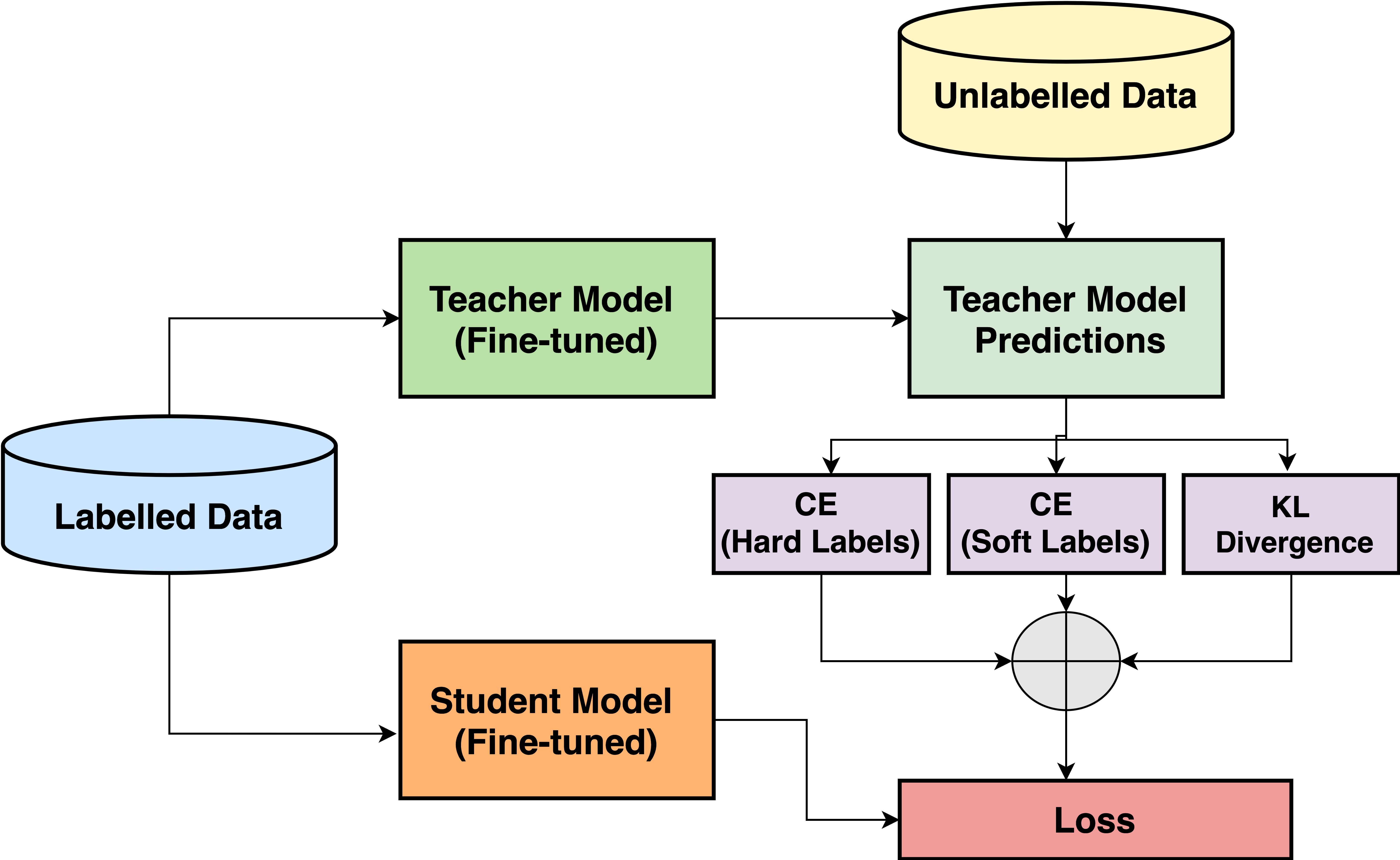}
\caption{High level overview of the process of Knowledge Distillation using Semi-Supervision}
\label{fig:block-diagram}
\end{figure}

Our approach is focused on leveraging knowledge distillation to build a highly accurate classifier with reduced cost of training and deployment. We train a model using manually labeled data, and call it as the ``Teacher''. We then pass the unlabelled data through this Teacher model. The output of the Teacher model is used as the label for this unlabelled data. This semi-supervision approach is used for data augmentation. Further, this augmented data is used to train a ``Student'' model. With the weakly labeled data, we employ multiple strategies to improve the performance of the ``Student'' model. Figure~\ref{fig:block-diagram} shows a high level system overview.
 
\subsection{Cross entropy as the loss function with hard labels}
\label{ce-hard-labels}
In this approach, we leverage the teacher model to obtain weak labels for unlabelled data. For a given sample we assign the most confident prediction (from the ``Teacher'') as its label. In other words, the output probabilities for each sample are converted into one hot encodings considering the prediction with highest probability as the true label. 

\subsection{Cross Entropy as the loss function with soft labels}
\label{CE_soft_labels}
Similar to the previous approach, we leverage the teacher model to obtain weak labels on a dataset. But instead of one hot encodings, we consider the probability distribution of the predictions as the labels for the samples. In other words, we perform semi supervision while ensuring to replicate the teacher's behaviour when labelling the samples. 
\subsection{KL Divergence as the loss function with soft labels}
Inspired by the results of Section~\ref{CE_soft_labels}, we leverage the probability distribution based labels but make use of 
KL divergence~\cite{kullback1997information} loss instead of Cross Entropy loss.  
Results show that this strategy performs the best to learn the difference between two probability distributions (the predictions from the student and the teacher models).



\section{Experiments}
\label{Experiments}
\subsection{Dataset}
Table~\ref{tab:my-table-data} shows the different datasets considered for the experiments. The initial training data comprises of 41,539 customer transactions sampled from September to December (2019) time frame - each of these transactions had an associated customer message. The free text message of every transaction was annotated by a team of three SMEs, and mapped to one of the ten pre-defined categories.  The list of categories considered are as follows:  \{'Food', 'Grocery', 'Package', 'Medicines', 'Household Items', 'Cigarettes', 'Clothes', 'Electronics', 'Keys', 'Documents/Books'\}.

Additionally, we consider 285,235 unlabelled customer transactions sampled from January to April (2020) for the semi-supervision experiments. For benchmarking the performance of different classification approaches, we label 20,156 customer transactions from April to construct a test dataset. This test set from April, containing 20,156 samples are \textbf{not} used for the semi-supervision experiments.  

\begin{table}[ht]{
\begin{center}
\begin{tabular}{|l|l|l|}
\hline
\textbf{Dataset} & \textbf{Duration} & \textbf{Size} \\ \hline
     Train      &  Sept - Dec & 41,539                  \\ \hline
     Unlabelled & Jan - Apr.  & 285,235                 \\ \hline
     Test       & April       & 20,156                  \\ \hline
\end{tabular}%
\end{center}
}
\caption{Dataset description}
\label{tab:my-table-data}
\end{table}
\subsection{Training the Teacher model}
In the first step, we train multiple models using the {\em Train} dataset and test on {\em Test} dataset to identify the candidate teacher model for our Knowledge Distillation experiments. For the purpose of the experiments, we consider XgBoost~\cite{chen2015xgboost, pedregosa2011scikit}, BiLSTM~\cite{gers1999learning, abadi2016tensorflow}, ALBERT~\cite{lan2019albert} and RoBERTa~\cite{liu2019roberta}. Table~\ref{table:pre-distill} shows the F1-scores for different models considered for this experiment. We observe that ALBERT and RoBERTa outperform BiLSTM and XgBoost. Therefore, RoBERTa is chosen as the teacher model for the next set of experiments. 
\begin{table}[htbp]{
\begin{center}
\begin{tabular}{|l|l|l|}
\hline
\textbf{Model} & \textbf{F1 Score} & \textbf{Accuracy} \\ \hline
XgBoost        &    0.60            &           63        \\ \hline
BiLSTM         &    0.65            &           73        \\ \hline
ALBERT         &    0.70            &           78       \\ \hline
RoBERTa        &    0.74            &           82       \\ \hline
\end{tabular}
\end{center}}
\caption{F1-scores and Accuracies for different classification models trained on Train dataset and tested on Test dataset}
\label{table:pre-distill}
\end{table}
\subsection{Generate Weakly Labeled data using Semi-Supervision}
\label{Generate Weakly Labeled data}
In the second step, we leverage the teacher model described in the previous subsection to extract weakly labeled samples for the {\em Unlabelled} dataset to augment the training dataset. To reduce the probability of selecting mislabeled samples, we set an empirical threshold of 95\% confidence in the label prediction as the criteria to accept a sample into the pool of training samples. At the end of this process, we obtain 93,820 additional training samples (\textit{weakly labelled by the teacher}).

\subsection{Leverage Knowledge Distillation to train a Student model}
Based on the production bottlenecks on number of parameters, the ALBERT(base) with \~11 million parameters  was chosen as the student model from the set of SOTA models. The model details can be found in Table~\ref{tab:parameter-comparison} under Appendix~\ref{appendix:settings}.

The data from Section~\ref{Generate Weakly Labeled data} is used further to ``teach'' the student models' making use of the 3  strategies mentioned in Section~\ref{Methodology}. The student model ALBERT is already trained (fine-tuned) on the labelled {\em Train} dataset which was used for training the teacher. R-ALBERT, the new student model based on the ALBERT architecture ends up performing the best and even better than the ``teacher'' model on our {\em Test} dataset. Similar pattern was observed in~\cite{eisenschlos2019multifit}.

\subsection{Reproducibility}\label{reproduce}
In order to validate the reproduciblity of our approach, we ran similar experiments on the 20Newsgroup~\cite{rennie20newsgroups} dataset.
\section{Results}
As shown in Table~\ref{tab:post-distill}, the Student model tends to perform better than its base version (the model which did not have a Teacher). We validate the statistical significance of the performance improvement using Stuart Maxwell Test~\cite{stuart1955test, maxwell1970comparing}. As shown in Table~\ref{tab:stuart-test-table}, the performance improvement over the base model is significant. Moreover, we observe that our approach achieved similar performance when compared to human annotated data, despite the change in data distributions and textual patterns. Also, from Table~\ref{tab:Reproducibility -newsgroup-20} we observe that the given method is reproducible on the 20Newsgroup dataset.Further, we observe that the distillation over RoBERTa's predictions gave an improvement of 8\% when compared to fine-tuning directly on part of the labelled dataset.





\begin{table}[t]
{%
\begin{center}
\begin{tabular}{|l|l|l|}
\hline
\textbf{Model} & \textbf{F1 Score} & \textbf{Accuracy} \\ \hline
R-ALBERT - OHE      &     0.72      &           83        \\ \hline
R-ALBERT - CE       &     0.65      &           64        \\ \hline
R-ALBERT - KL       &     0.73      &           84       \\ \hline
RoBERTa             &     0.18      &           40       \\ \hline
\end{tabular}%
\end{center}
}
\caption{ Comparison of F1-scores on internal benchmark using different approaches}
\label{tab:post-distill}
\end{table}

\begin{table}[]
\centering
\begin{tabular}{|l|l|l|}
\hline
\textbf{Model} & \textbf{F1-score} & \textbf{Accuracy (\%)} \\ \hline
ALBERT         & 0.63              & 65                     \\ \hline
R-ALBERT-KL    & 0.70              & 73                     \\ \hline
RoBERTa        & 0.88              & 87                     \\ \hline
\end{tabular}
\caption{F1-scores on the 20Newsgroup dataset with 8,073 train samples, 7,037 weakly labeled samples (after 95\% threshold) and 805 samples}
\label{tab:Reproducibility -newsgroup-20}
\end{table}


\section{Conclusion}\label{Conclusion}
We explore a generalised distillation framework on transformer based architecture which shows that ``Students'' can be made better with the help of the weak labels generated by a good ``Teacher''.  Given the focus on having smaller models and to ensure that we make effective utilisation of resources, we feel this distillation method will pave path for more research in the future. This framework can also help reduce manual labelling efforts.

\bibliography{anthology,eacl2021}
\bibliographystyle{acl_natbib}

\clearpage
\newpage
\appendix
\section{Appendix}
\label{appendix:settings}

\begin{table*}[!htbp]{%
\centering
\begin{tabular}{|l|l|l|l|l|}
\hline
\textbf{Model 1} & \textbf{Model 2} & \textbf{Chi-square} & \textbf{DoF} & \textbf{p-value (\textless{})} \\ \hline
ALBERT      & RoBERTa      & 2185.71 & 9 & 2.2e-16 \\ \hline
R-ALBERT-KL & R-ALBERT-OHE & 955.61  & 9 & 2.2e-16 \\ \hline
R-ALBERT-KL & RoBERTa      & 955.61  & 9 & 2.2e-16 \\ \hline
\end{tabular}
}
\caption{The results of Stuart Maxwell Test's can be found, where we compare different models and we can see that the increase/decrease in performance is statistically significant.}
\label{tab:stuart-test-table}
\end{table*}

\begin{table}[!htbp]
\centering
\begin{tabular}{|l|l|}
\hline
\textbf{Model}      & \textbf{Parameters (in millions)} \\ \hline
ALBERT(base)        & 11                                \\ \hline
distilBERT(base)    & 66                                \\ \hline
distilRoBERTa(base) & 82                                \\ \hline
RoBERTa(base)       & 125                               \\ \hline
RoBERTa(large)      & 355                               \\ \hline
\end{tabular}
\caption{As seen from the table, ALBERT(base) has the least no of parameters when compared to the rest of the models. hence we chose ALBERT as the student. distilRoBERTa would have been a good choice, given that tit has signals from RoBERTa, but due to memory requirements in production we choose ALBERT(base)}
\label{tab:parameter-comparison}
\end{table}
\end{document}